\documentclass[letterpaper, 10 pt, conference]{ieeeconf}  

\IEEEoverridecommandlockouts                              

\usepackage{algpseudocode}

\usepackage{graphicx}
\usepackage{algorithm}
\usepackage{xcolor}
\usepackage{booktabs}
\usepackage{multirow}
\usepackage{balance}
\usepackage{amsmath}
\usepackage{bm}
\usepackage{subcaption}
\usepackage[font=small,labelfont=small]{caption}
\usepackage{amssymb}
\usepackage{xcolor}
\usepackage{tikz}
\definecolor{samtarget}{HTML}{82B0D2}
\definecolor{samrobot}{HTML}{FA7F6F}

\title{Foundation and Small Models Coordination for Visuomotor Policy Learning}

\author{
Haoran Ding$^{1}$, Liang Ma$^{1}$, Yaxun Yang$^{1}$, Wen Yang$^{1}$, 
Tianyu Liu$^{1}$,\\ Xiaodan Liang$^{1}$,
Dezhen Song$^{1}$, Ivan Laptev$^{1}$, Yoshihiko Nakamura$^{1}$, 
Anqing Duan$^{1}$
\thanks{
$^{1}$Authors are at Mohamed bin Zayed University of Artificial Intelligence (MBZUAI), Abu Dhabi, UAE.
Email: {\tt\small \{firstname.lastname\}@mbzuai.ac.ae}
}
}

\begin{document}
\maketitle
\thispagestyle{empty}
\pagestyle{empty}

\begin{abstract}
Visuomotor policy learning enables robots to perform a wide range of tasks, but small policy models often remain sensitive to changes in object and background appearance. In this work, we investigate the coordination of pretrained vision foundation models with small policy models to improve appearance generalization. We propose a framework in which a small policy model operates on task-relevant visual observations constructed through semantic repainting. A segmentation foundation model identifies the robot and target object, which are rendered with fixed role colors on a constant background. An alternative representation replaces the target’s role color with normalized monocular depth predicted by a depth foundation model, providing additional geometric cues. The perception models are adapted using in-distribution data where needed and held fixed during policy training. This design combines the perceptual capabilities of foundation models with a small policy model trained on the resulting observations for action prediction. Evaluations with flow matching policies on simulation benchmarks, together with experiments on two real-world robotic tasks, demonstrate substantial improvements in task success under the evaluated appearance shifts.
\end{abstract}

\section{Introduction}


Small visuomotor policy models can learn effective manipulation skills from demonstrations~\cite{chi2025diffusion, zhang2025flowpolicy}, yet their performance can decline when visual conditions differ from training~\cite{garcia2023robust, PumacayRSS24}. Changes in object color, background appearance, or clutter may disrupt execution even when the task remains unchanged. Large-scale robot policies and vision--language(-action) systems support a broader range of tasks and settings~\cite{brohan2022rt, kim2024openvla, li2023vision}, but their capabilities rely on large robot datasets and substantial pretraining~\cite{team2024octo, o2024open}, making further policy scaling costly in data and computation.

Our focus is appearance generalization within the same task, where the task objective and required manipulation skill remain unchanged. Vision foundation models pretrained on broad visual data offer reusable capabilities for identifying and segmenting task-relevant entities. This motivates using their perceptual knowledge to support action learning in small policy models. The central question is how to design a perception--policy interface that reduces the influence of appearance variation, allowing small policy models to execute learned tasks reliably across changes in object and background appearance.

\begin{figure}[t]
  \centering
  \includegraphics[width=\linewidth]{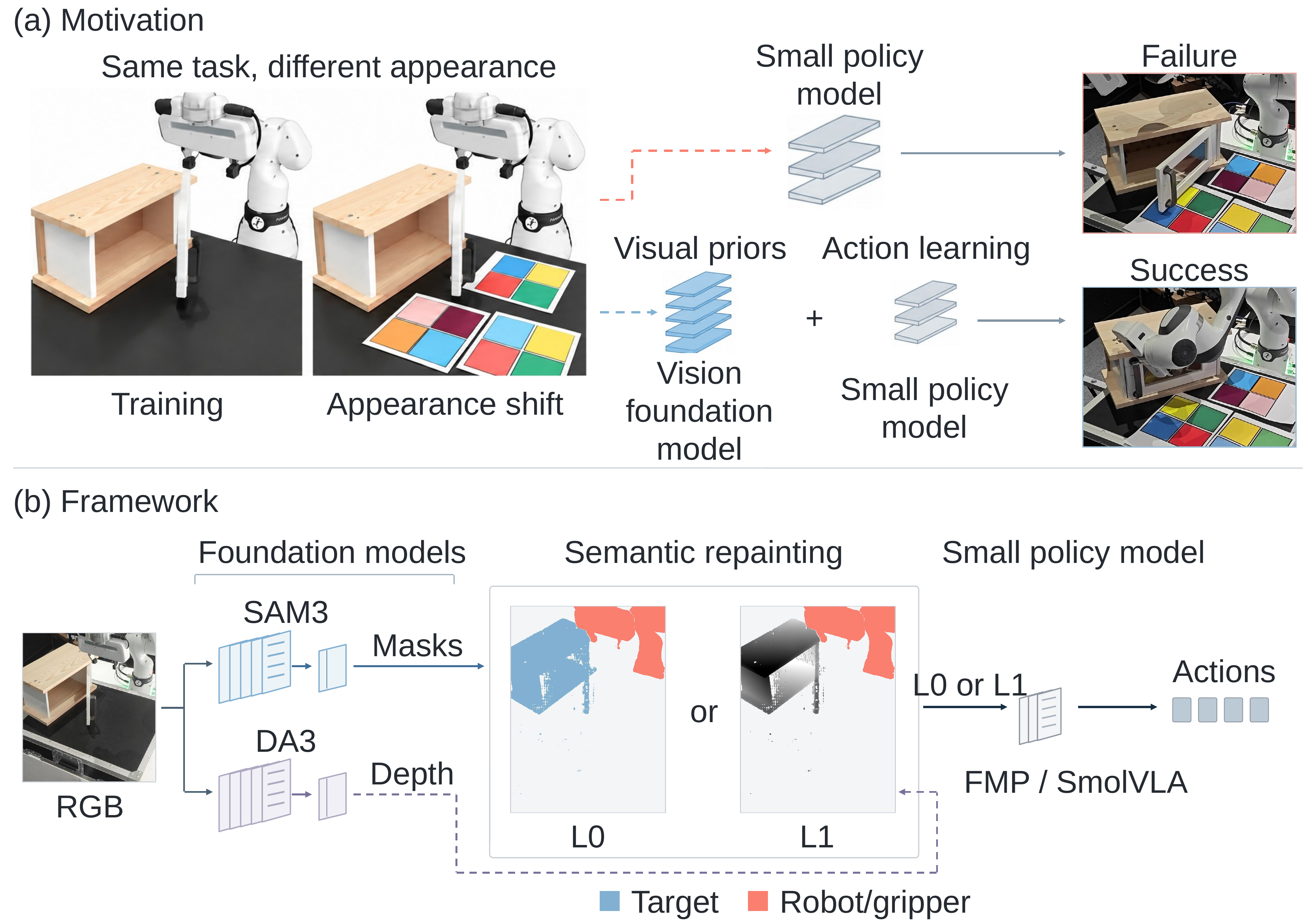}
\caption{Coordinating vision foundation models and small policy models.
(a) Motivation under appearance shifts.
(b) Task-relevant visual representations connect foundation-model perception with action learning in small policy models.}
  \label{fig:method}
\end{figure}

To this end, we construct an explicit perception--policy interface, as illustrated in Fig.~\ref{fig:method}. Promptable segmentation foundation models~\cite{ravi2024sam2, carion2025sam3} identify task-specified entities, and semantic repainting converts their masks into visual observations for the policy. The perception models are adapted using in-distribution (ID) data where needed, then held fixed while the policy learns actions from these observations.

The interface provides two observation variants. L0 renders the robot or gripper and target object with fixed role colors on a constant background, suppressing their original colors and textures as well as background appearance. L1 replaces the target-object region with monocular depth from Depth Anything~3, normalized within that region, to provide additional geometric cues. Both variants remain standard three-channel images and can be used for policy training without modifying the policy architecture.

Our contributions are as follows:
\begin{itemize}
    \item We develop a semantic--geometric visual interface
    that coordinates pretrained vision foundation models
    with small policy models, separating perception adaptation
    from action learning without changes to the policy architecture.

    \item We analyze perception adaptation and interface
    design, identifying the importance of robot/gripper
    information, the benefits of appearance canonicalization,
    and the task-dependent value of depth in the evaluated
    settings.

    \item We demonstrate appearance-robustness gains with a Flow Matching Policy (FMP)
    across simulation benchmarks~\cite{robomimic2021,
    taomaniskill3, james2019rlbench} and two real-world Franka
    tasks, complemented by SmolVLA evaluations on RLBench.
\end{itemize}

\section{Related Work}
We discuss visuomotor policy learning and how pretrained vision models are connected to action prediction.

\subsection{Visuomotor Policy Learning}
Visuomotor policies learn actions from visual observations, often combined with proprioception, language, or goal images~\cite{jiang2022vima, shridhar2023perceiver, jang2022bc, mees2022calvin}. Actions can be predicted directly~\cite{li2023vision, brohan2022rt} or generated through diffusion and flow matching~\cite{chi2025diffusion, ding2025fast, pf2mp}. Vision--language conditioning is compatible with token-based~\cite{brohan2022rt, zitkovich2023rt, kim2024openvla}, regression-based~\cite{li2023vision}, and diffusion/flow-based action decoders~\cite{intelligence2025pi, li2024cogact, liu2025hybridvla}. Models such as RT-2 and OpenVLA transfer pretrained vision--language representations into action prediction. Our framework adds a separate perceptual stage that constructs task-relevant image observations for the downstream policy.

Camera configurations in RoboMimic and RLBench~\cite{robomimic2021, james2019rlbench} and synthesized wrist views~\cite{ding2025imagination} determine the visual information available to a policy. At a larger scale, Open X-Embodiment~\cite{o2024open} supports broader policy pretraining in systems such as Octo and OpenVLA~\cite{team2024octo, kim2024openvla}. Such large-scale policy pretraining requires substantial robot data and computation. We investigate a complementary use of pretrained capabilities: vision foundation models construct observations that suppress appearance variation, supporting robust action learning while retaining a small downstream policy model and its architecture.

\subsection{Vision Foundation Models for Policy Learning}
SAM series work~\cite{kirillov2023segment, ravi2024sam2, carion2025sam3} provide reusable segmentation capabilities that can support robot policies through different interfaces. SAM2Act integrates multi-resolution visual features into a multi-view policy~\cite{fang2025sam2act}, while SAM2Grasp connects frozen SAM2 features to a lightweight action head~\cite{wu2025sam2grasp}. ImitDiff uses semantic masks to guide visual feature fusion~\cite{dong2025imitdiff}. We similarly draw on pretrained perception, but pass its outputs to the policy as rendered images processed by the policy's existing visual encoder.

The purpose of the perceptual representation also differs across studies. Synthetic semantic augmentation expands training appearance diversity~\cite{yu2023scaling}, whereas our interface suppresses appearance variation during both training and inference. SHADOW uses robot masks to support cross-embodiment transfer~\cite{lepert2025shadow}; we use robot and target masks to reduce sensitivity to object and background appearance. OBEYED-VLA uses object-centric and geometric grounding to support VLA policies in clutter~\cite{vo2025clutter}. We share this use of external perceptual guidance and examine its connection to task-specific and compact pretrained policy models.

Explicit image-based interfaces are particularly close to our design. ARRO retains foreground RGB and introduces virtual guides~\cite{zhu2025arro}; our fixed role colors additionally remove the original colors and textures of retained entities. S$^{2}$-Diffusion concatenates semantic masks and monocular depth for category-level skill generalization~\cite{s2diffusion}. Our depth variant encodes target geometry in a three-channel image while retaining the robot's role color, enabling direct use of the existing policy encoder.

PEEK uses VLM-predicted paths and masking points to guide policies through modified images~\cite{zhang2025peek}. We share its division of perceptual guidance and action learning, but supply entity and geometry observations without generating action-path guidance.

Within this line of research, we investigate coordination between vision foundation models and small policy models through an explicit visual interface. Our focus is how task-relevant observations constructed by foundation models support robust action learning under object and background appearance changes, with perception adaptation and policy learning performed in separate stages.

\section{Methodology}

As illustrated in Fig.~\ref{fig:method}, we coordinate vision foundation models and small visuomotor policy models through an explicit visual interface. A segmentation foundation model identifies task-relevant entities, and semantic repainting converts its predictions into observations for policy learning. For L1, a depth foundation model supplies additional geometric cues. The small policy model learns to predict actions from these observations, aiming to reduce sensitivity to object and background appearance changes.

\subsection{Problem Setup and Model Coordination}

We study visuomotor policy learning from a dataset of expert demonstrations
$\mathcal{D}=\{\tau^{(i)}\}_{i=1}^{N}$, where each trajectory contains RGB observations $I_t \in \mathbb{R}^{H \times W \times 3}$ and expert action targets $a_t \in \mathbb{R}^{d_a}$.
Given a task specification $c$, we construct the policy's visual input as
\begin{equation}
\tilde{o}_t = g_{\ell}(I_t,c),
\qquad
\ell \in \{\mathrm{L0},\mathrm{L1}\},
\label{eq:obs_extraction}
\end{equation}
where L0 uses semantic role colors, and L1 additionally incorporates target-object depth. Each policy is trained and evaluated with a fixed choice of representation. 

Perception adaptation and policy learning are performed separately. Where needed, the vision foundation models are adapted using only in-distribution (ID) training data, as detailed in Sec.~\ref{sec:perception_finetune}. Their parameters are then held fixed during policy training and evaluation. The policy learns from the constructed observations and expert actions, without propagating its training objective into the perception models.

\subsection{Task-Relevant Perception and Semantic Repainting (L0)}
\label{sec:obs_extraction}

We use SAM3~\cite{carion2025sam3} to obtain task-relevant masks from the raw RGB image. The task specification $c$ determines text prompts $p^{\mathrm{r}}$ and $p^{\mathrm{o}}$ for the robot or gripper and the target object, respectively, using a fixed prompt template per benchmark:
\begin{equation}
\begin{aligned}
M_t^{\mathrm{r}} &= \mathrm{SAM3}(I_t,p^{\mathrm{r}}),\hspace{0.4em}
M_t^{\mathrm{o}} = \mathrm{SAM3}(I_t,p^{\mathrm{o}}),
\end{aligned}
\label{eq:sam3_masks}
\end{equation}
where $M_t^{\mathrm{r}},M_t^{\mathrm{o}}
\in \{0,1\}^{H \times W}$.
When a prompt produces multiple instance masks, we take their union.

The masks are converted into a three-channel observation through semantic repainting:
\begin{equation}
I_t^{\mathrm{L0}}
=
\mathcal{R}\!\left(
M_t^{\mathrm{r}},M_t^{\mathrm{o}};
\mathbf{c}_{\mathrm{r}},\mathbf{c}_{\mathrm{o}},\kappa
\right),
\label{eq:l0_repaint}
\end{equation}
where $\mathcal{R}$ is a semantic rendering operator that fills the robot and target regions with specified RGB colors or per-pixel three-channel values, and assigns a constant color $\kappa$ to the background. For L0, $\mathbf{c}_{\mathrm{r}}$ and $\mathbf{c}_{\mathrm{o}}$ are fixed role colors, and we use a black background, $\kappa=\mathbf{0}$.

L0 preserves the shapes and positions of the segmented regions in the image while removing their original colors and textures and suppressing background appearance. For the same masks, the rendered observation does not depend on the original RGB appearance. This gives the small policy model spatial and semantic information for action learning.

\subsection{Geometric Target Representation (L1)}
\label{sec:depth_injection}

L0 assigns a uniform color to each segmented region, removing visual variation within that region. To provide additional geometric information, L1 incorporates monocular depth predicted by Depth Anything 3 (DA3)~\cite{depthanything3}:
\begin{equation}
\hat{D}_t = \mathrm{DA3}(I_t),
\qquad
\hat{D}_t \in \mathbb{R}^{H \times W}.
\end{equation}
Depth is estimated from the original RGB image and normalized within the target-object mask.

Let
$\Omega_t^{\mathrm{o}}
=\{(u,v)\mid M_t^{\mathrm{o}}(u,v)=1\}$
denote the target region. For pixels in this region, we compute
\begin{equation}
D_t^{\mathrm{norm}}(u,v)
=
\frac{\hat{D}_t(u,v)-d_{\min}}
     {d_{\max}-d_{\min}+\epsilon},
\qquad
(u,v)\in\Omega_t^{\mathrm{o}},
\label{eq:depth_norm}
\end{equation}
where $d_{\min}$ and $d_{\max}$ are the minimum and maximum predicted depths within $\Omega_t^{\mathrm{o}}$, and $\epsilon>0$ stabilizes the denominator. We set $D_t^{\mathrm{norm}}$ to zero outside the target region. This normalization represents relative depth variation within the target object.

We repeat the normalized depth map across three channels,
$I_t^{\mathrm{dep}}=\mathrm{tile}(D_t^{\mathrm{norm}},3)$,
and use it as the target-region fill:
\begin{equation}
I_t^{\mathrm{L1}}
=
\mathcal{R}\!\left(
M_t^{\mathrm{r}},M_t^{\mathrm{o}};
\mathbf{c}_{\mathrm{r}},I_t^{\mathrm{dep}},\kappa
\right).
\label{eq:depth_overwrite}
\end{equation}
L1 uses the same semantic rendering rule as L0, replacing the target's role color with normalized depth while retaining the robot's role color and the constant background. Both variants remain three-channel images compatible with the policy's existing visual encoder.

Algorithm~\ref{alg:interface} summarizes the observation construction used for policy training and deployment.

\begin{algorithm}[t]
\caption{Observation Construction for Small Policy Model}
\label{alg:interface}
\small
\algrenewcommand\algorithmicindent{1em}
\begin{algorithmic}[1]
\Require RGB image $I_t$, task specification $c$,
representation $\ell \in \{\mathrm{L0},\mathrm{L1}\}$,
fixed perception models and rendering colors
\Ensure Policy observation $\tilde{o}_t$

\State Form prompts $(p^{\mathrm{r}},p^{\mathrm{o}})$ from $c$
\State Obtain masks $(M_t^{\mathrm{r}},M_t^{\mathrm{o}})$
using Eq.~\eqref{eq:sam3_masks}
\State Union instance masks within each semantic role

\If{$\ell=\mathrm{L1}$}
    \State Estimate depth:
        $\hat{D}_t \gets \mathrm{DA3}(I_t)$
    \State Normalize target depth using Eq.~\eqref{eq:depth_norm}
    \State Tile depth:
        $I_t^{\mathrm{dep}}
        \gets \mathrm{tile}(D_t^{\mathrm{norm}},3)$
    \State Render L1:
        $\tilde{o}_t \gets I_t^{\mathrm{L1}}$
        using Eq.~\eqref{eq:depth_overwrite}
\Else
    \State Render L0:
        $\tilde{o}_t \gets I_t^{\mathrm{L0}}$
        using Eq.~\eqref{eq:l0_repaint}
\EndIf

\State \Return $\tilde{o}_t$
\end{algorithmic}
\end{algorithm}

\subsection{Learning Small Visuomotor Policies}
\label{sec:policy}

We primarily instantiate the downstream policy with a Flow Matching Policy (FMP)~\cite{braun2024riemannian,rouxel2024flow}, retaining its visual encoder, 1D U-Net, and action parameterization. Let $\mathcal{D}_{\ell}$ denote the demonstration pairs obtained by replacing each RGB observation with $g_{\ell}(I_t,c)$. The policy is trained on these transformed observations and their associated expert action targets.

Conditioned on $\tilde{o}_t$, FMP learns a vector field
$v_{\theta}(x,s;\tilde{o}_t)$
that transports samples from a base distribution $p_0$ to the conditional action distribution. Here, $t$ indexes environment time and $s \in [0,1]$ denotes flow time. We sample $(\tilde{o}_t,a_t)\sim\mathcal{D}_{\ell}$, $x_0\sim p_0$, and $s\sim\mathcal{U}(0,1)$, and minimize
\begin{equation}
\mathcal{L}_{\mathrm{FM}}(\theta)
=
\mathbb{E}\!\left[
\left\|
v_{\theta}(x_s,s;\tilde{o}_t)-\dot{x}_s
\right\|_2^2
\right],
\label{eq:fm_loss}
\end{equation}
where $x_s$ follows the training probability path from $x_0$ to the expert action target $a_t$, and $\dot{x}_s$ is its target velocity. At inference, integrating the learned vector field from $s=0$ to $s=1$ produces an action prediction.

We additionally evaluate this visual interface with SmolVLA~\cite{shukor2025smolvla} on RLBench. SmolVLA is fine-tuned on the selected observation representation while retaining its policy architecture and training objective. This evaluates the same perception--policy interface with a different downstream model.

During deployment, the fixed perception models construct the selected L0 or L1 observations from incoming RGB images, and the trained policy predicts actions from these observations. The perception models and policy remain fixed under the evaluated appearance shifts.

\section{Experiments}

We evaluate whether coordinating vision foundation models with small policy models improves task success under appearance shifts. Our experiments compare policies learned from raw RGB with policies learned from task-relevant observations constructed by the foundation models, assessing performance under both in-distribution (ID) and held-out visual conditions. We primarily use a Flow Matching Policy (FMP) and additionally evaluate SmolVLA~\cite{shukor2025smolvla} on RLBench to examine the same perception--policy interface with a second policy backbone.

\subsection{Experimental Setup and Evaluation Protocol}
\label{sec:exp_setup}

\begin{figure*}[t]
  \centering
  \captionsetup[subfigure]{skip=3pt}

  \begin{subfigure}[t]{0.75\textwidth}
    \centering
    \vspace{0pt}
    \includegraphics[width=\linewidth]{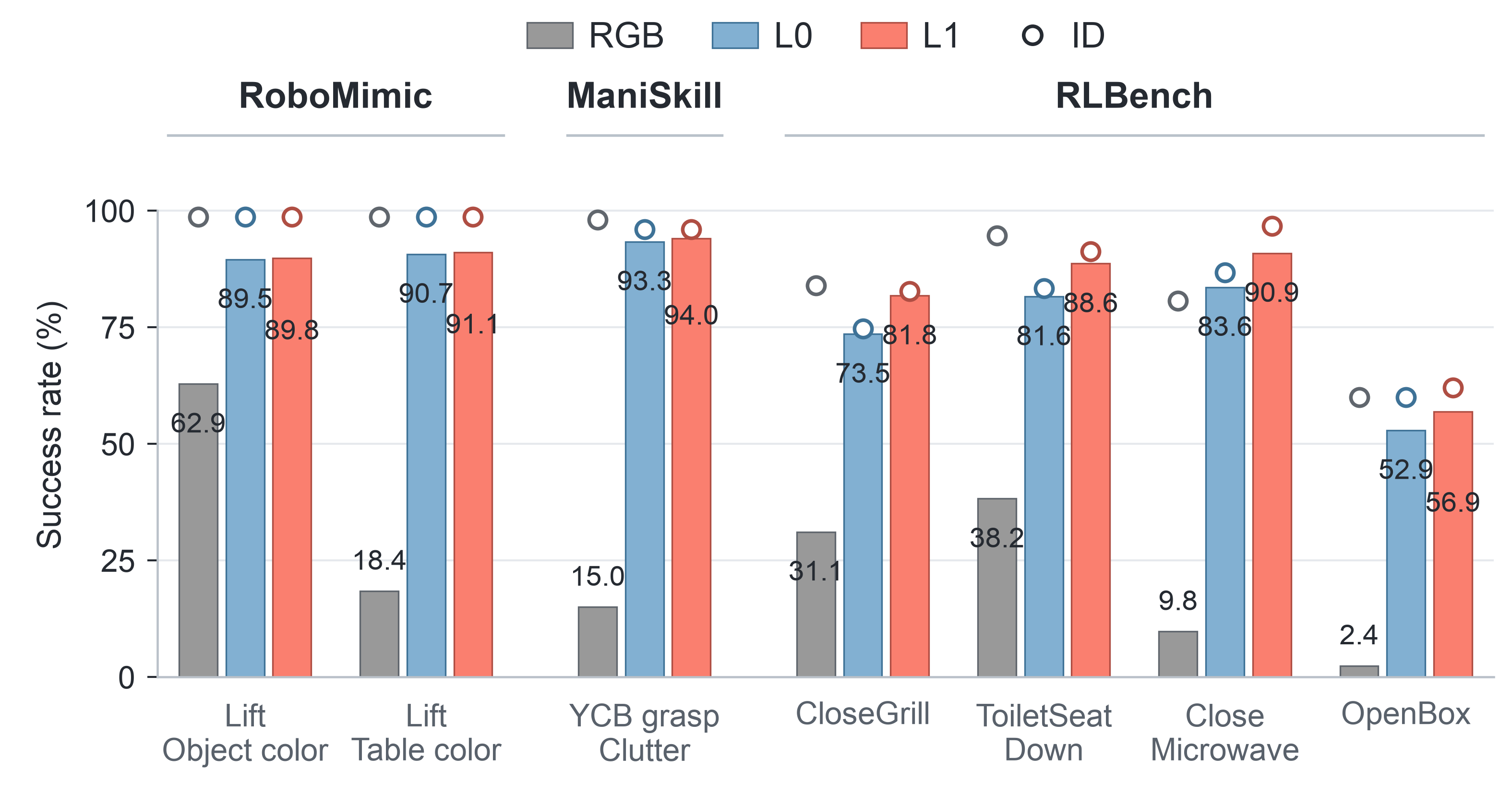}
    \caption{}
    \label{fig:sim_robustness}
  \end{subfigure}
  \hfill
  \begin{subfigure}[t]{0.23\textwidth}
    \centering
    \vspace{0pt}
    \includegraphics[width=\linewidth]{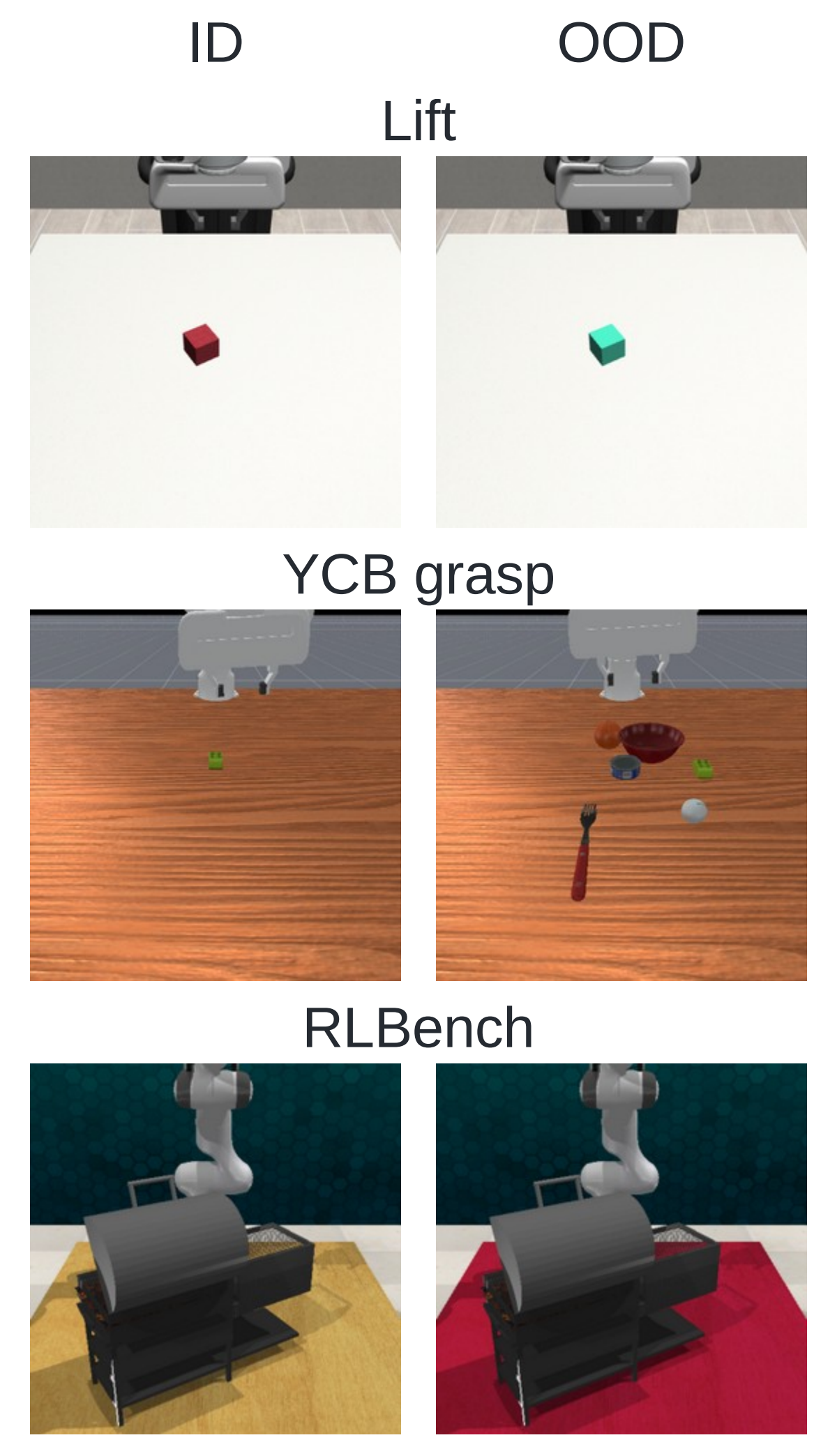}
    \caption{}
    \label{fig:exp_envs}
  \end{subfigure}
    \caption{Simulation evaluation under appearance shifts. (a) FMP performance: bars show mean success over three held-out conditions (OOD1--3); open circles indicate ID success. (b) ID and one representative OOD condition for each illustrated task.}
  \label{fig:sim_evaluation}
\end{figure*}


\paragraph{Benchmarks and tasks}
We evaluate on RoboMimic (Lift)~\cite{robomimic2021}, ManiSkill (YCB object grasp)~\cite{taomaniskill3}, and RLBench (CloseGrill, ToiletSeatDown, CloseMicrowave, and OpenBox)~\cite{james2019rlbench}. We also evaluate two tasks on a real Franka robot, as described in Sec.~\ref{sec:real_robot}.

\paragraph{Observation variants}
We compare three visual inputs to the policy:
(i) RGB, the original color observations;
(ii) L0, semantic repainting based on segmentation foundation-model predictions, which renders the robot/gripper and target object with fixed role colors on a constant background (Sec.~\ref{sec:obs_extraction});
and (iii) L1, which incorporates geometry from a depth foundation model by replacing only the target-object region with normalized monocular depth, while retaining the robot/gripper color and constant background (Sec.~\ref{sec:depth_injection}).

Within each task and policy backbone, all variants use the same demonstrations, policy architecture, visual encoder, optimization settings, and policy training budget. Each policy is trained or fine-tuned on its selected observation representation. This controlled comparison evaluates the effect of the constructed visual interface on action learning and robustness.

\paragraph{Appearance-shift protocol}
ID evaluation uses the training visual setting. OOD evaluation introduces appearance conditions held out from both perception adaptation and policy learning. We consider two types of shifts: OOD-Obj changes the target object's color, while OOD-Bg changes background or support-surface colors and/or introduces clutter distractors. Task objectives and success criteria remain unchanged across conditions. Fig.~\ref{fig:sim_evaluation} presents FMP performance in panel (\subref{fig:sim_robustness}) and representative ID and OOD environments in panel (\subref{fig:exp_envs}).

\paragraph{Implementation details} SAM3 receives task-specific text prompts identifying the robot/gripper and target object. For L1, Depth Anything 3 (DA3) estimates depth from the original RGB observation. The estimated depth is normalized within the target-object mask before being incorporated into the policy input. Where adaptation is needed, the perception models are adapted using only ID training data. Their parameters are then held fixed during policy learning and evaluation, separating perception adaptation from action learning. No further perception or policy adaptation is performed under OOD conditions. We report success rates averaged over three random seeds.

\subsection{Foundation Model Adaptation and Usage}
\label{sec:perception_finetune}

We adapt the perception models using LoRA implemented with PEFT~\cite{peft}, with ID-only supervision: reference masks for SAM3 and paired RGB--depth data for DA3. DA3 adaptation is limited to simulation; real-robot experiments use pretrained DA3. During adaptation, pretrained parameters are frozen and only LoRA parameters are optimized, with rank $r=8$, $\alpha=16$, and dropout of 0.05. All perception parameters remain fixed during policy training and evaluation.

\paragraph{SAM3 adaptation}
We insert LoRA adapters into the query and value projections of the self-attention layers in the final four transformer blocks of the SAM3 text encoder. Each RGB frame provides two training samples: a robot/gripper union mask and a target-object union mask, each paired with a fixed text prompt.

We train for 50 epochs using AdamW with a learning rate of $5 \times 10^{-4}$ and weight decay of $10^{-2}$ under mixed precision. The segmentation objective combines binary cross-entropy and Dice losses with weights of 1 and 0.5, respectively. Within each benchmark, the same adapted SAM3 checkpoint is shared across policy runs.

\paragraph{DA3 adaptation}
We insert LoRA adapters into selected encoder projections in DA3, including attention query, key, value, and output projections and MLP input and output projections. The adapted model supplies depth estimates for the L1 observation.

RGB inputs follow DA3 preprocessing: upper-bound resizing to a target resolution of 504 pixels, spatial dimensions aligned to multiples of the patch size of 14, and normalization using ImageNet statistics. Using paired RGB--depth supervision, we optimize the scale-invariant SILog loss with coefficient $\lambda=0.85$ for 100 epochs using AdamW, with a learning rate of $10^{-4}$ and zero weight decay.

\subsection{Simulation Benchmarks}
\label{sec:sim_results}

We use FMP to evaluate how observations constructed by vision foundation models support small policy models across RoboMimic, ManiSkill, and RLBench. Comparisons follow the controlled setup in Sec.~\ref{sec:exp_setup}. Fig.~\ref{fig:sim_evaluation}(\subref{fig:sim_robustness}) reports ID success and mean OOD success across three held-out appearance conditions. Unless otherwise specified, OOD results below refer to this mean.

\paragraph{RoboMimic: object and tabletop color shifts}
On Lift, we vary the cube color (OOD-Obj) and tabletop color (OOD-Bg) separately while retaining the task dynamics. All three observation variants achieve 98.7\% ID success. Under appearance shifts, RGB success decreases to 62.9\% for object recoloring and 18.4\% for tabletop recoloring. L0 improves these results to 89.5\% and 90.7\%, respectively.

L1 achieves similar OOD success, reaching 89.8\% under object recoloring and 91.1\% under tabletop recoloring, without consistently outperforming L0 across individual conditions. These results indicate that semantic repainting provides an effective visual interface for FMP under the tested color shifts, with limited additional improvement from depth in this task.

\paragraph{ManiSkill: grasping under clutter}
We evaluate grasping of a green Lego object after training in an uncluttered environment. OOD conditions introduce distractor objects and changes in scene layout. RGB success decreases from 98.0\% in ID to 15.0\% under OOD conditions. L0 and L1 both achieve 96.0\% ID success and retain high OOD success of 93.3\% and 94.0\%, respectively.

The observations constructed by the foundation models allow the small policy model to maintain task performance across the tested cluttered scenes. As on Lift, adding depth provides only a small improvement in mean success over semantic repainting alone.

\paragraph{RLBench: tabletop color shifts}
We evaluate CloseGrill, ToiletSeatDown, CloseMicrowave, and OpenBox under the training tabletop appearance and three held-out tabletop colors. RGB performance decreases under appearance shifts on all four tasks. For example, success falls from 80.7\% in ID to 9.8\% in OOD on CloseMicrowave, and from 60.0\% to 2.4\% on OpenBox.

L0 improves mean OOD success to 73.5\% on CloseGrill, 81.6\% on ToiletSeatDown, 83.6\% on CloseMicrowave, and 52.9\% on OpenBox. L1 further improves these means to 81.8\%, 88.6\%, 90.9\%, and 56.9\%, respectively. On CloseMicrowave, adding depth also increases ID success from 86.7\% with L0 to 96.7\% with L1. On CloseGrill and ToiletSeatDown, L0 has lower ID success than RGB, while L1 reduces these gaps.

The benefit of target depth is task-dependent, with larger gains on CloseMicrowave and smaller changes on Lift and ManiSkill. On OpenBox, L1 improves over L0 in OOD1 and OOD2 but yields slightly lower success in OOD3 (51.3\% versus 52.0\%). Together with the lower ID success of L0 on some tasks, these results suggest a trade-off between suppressing appearance variation and preserving information for action prediction.

\subsection{Real Robot Experiments}
\label{sec:real_robot}

\begin{figure*}[t]
  \centering
  \includegraphics[width=\textwidth]{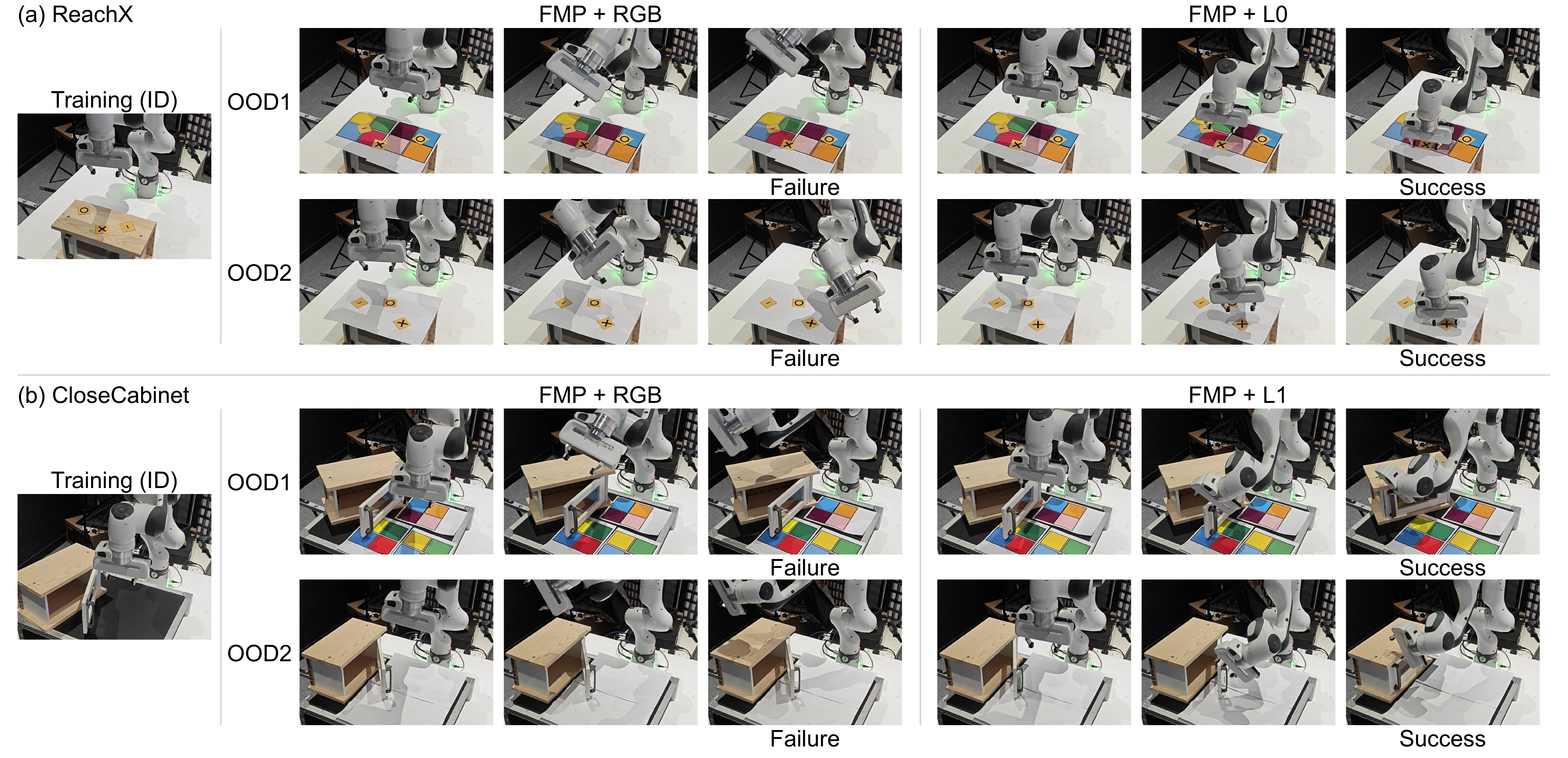}
  \caption{Real-robot rollouts under appearance shifts.}
  \label{fig:real_envs}
  \vspace{-0.6em}
\end{figure*}

\begin{table}[t]
\centering
\tiny
\setlength{\tabcolsep}{4.0pt}
\renewcommand{\arraystretch}{1.15}
\caption{Real-robot success rates (\%) under appearance shifts.
Ours uses L0 for ReachX and L1 for CloseCabinet.}
\label{tab:real_obs}
\resizebox{\linewidth}{!}{
\begin{tabular}{l|ccc|ccc}
\hline
& \multicolumn{3}{c|}{ReachX}
& \multicolumn{3}{c}{CloseCabinet} \\
\cline{2-7}
Obs. & ID & OOD1 & OOD2 & ID & OOD1 & OOD2 \\
\hline
RGB  & 61.7 & 21.7 & 25.0 & 85.0 & 45.0 & 48.3 \\
Ours & 88.3 & 83.3 & 81.7 & 81.7 & 78.3 & 75.0 \\
\hline
\end{tabular}
}
\vspace{-0.6em}
\end{table}

We evaluate the coordination of vision foundation models and small policy models on a real Franka robot. Two manipulation tasks examine whether policies learned from the constructed observations retain task success under changes in the surrounding visual appearance. Fig.~\ref{fig:real_envs} shows representative rollouts, and Table~\ref{tab:real_obs} reports the quantitative results.

\paragraph{Task 1: ReachX}
The robot moves its end-effector to touch the marker labeled X among three planar markers. Marker layout and support-surface appearance vary across the ID and OOD settings. We compare RGB with L0, using semantic observations to represent the robot and target for this planar reaching task.

\paragraph{Task 2: CloseCabinet}
The robot closes a partially open cabinet through contact with the door. We compare RGB with L1, which includes target-region depth to provide geometric cues for contact and spatial alignment.

\paragraph{Demonstrations and perception adaptation}
For each task, we collect 25 demonstrations using GELLO~\cite{gello}. Each training run uses 20 demonstrations, with the remaining five held out. We repeat training with three random splits.

SAM3 provides segmentation masks from text prompts for constructing the training observations. When text prompting produces an incorrect mask, a minimal click prompt is used to correct it. The resulting masks provide supervision for LoRA adaptation of SAM3 using the training data only. During deployment, SAM3 uses text prompts without click correction.

For L1, depth is estimated using the pretrained Depth Anything 3 model without adaptation on real-robot data. After perception preparation, the perception parameters are held fixed while the policy learns from the constructed observations. No OOD data is used for perception adaptation or policy training.

\paragraph{Deployment and evaluation}
All variants use FMP to predict target joint states from the current and previous visual observations. Each policy receives its selected RGB, L0, or L1 representation. The franky control library~\cite{franky} executes the predicted targets through low-level joint control.

We evaluate each trained model for 20 rollouts under each ID and OOD condition, varying the object or cabinet pose and support-surface appearance. Success rates are averaged over the three training runs. Neither the perception models nor the policy undergo further adaptation during evaluation.

On an RTX 4060 Ti, SAM3 inference takes approximately 15 ms per prompt, or 30 ms per frame for the robot and target prompts together. DA3 inference takes approximately 52 ms per frame. These perception runtimes are compatible with the target-joint tracking setup used for the two tasks.

\paragraph{Results}
On ReachX, RGB success decreases from 61.7\% in ID to 21.7\% and 25.0\% under OOD1 and OOD2. L0 achieves 88.3\% ID success and retains 83.3\% and 81.7\% success under the two OOD conditions. Semantic repainting therefore improves success in both the training visual setting and the held-out settings, while reducing the decline associated with appearance changes.
On CloseCabinet, RGB achieves 85.0\% ID success but decreases to 45.0\% and 48.3\% under OOD1 and OOD2. L1 achieves 81.7\% ID success and retains 78.3\% and 75.0\% under OOD. Although its ID success is lower than that of RGB, L1 exhibits a smaller decrease under the held-out appearances.

The representative rollouts in Fig.~\ref{fig:real_envs} illustrate RGB-policy failures and successful executions with the constructed observations under OOD conditions. These results support using fixed foundation-model perception to provide task-relevant observations for small policy models on a real robot, with improved robustness under the evaluated appearance shifts.

\subsection{Ablations and Additional Evaluations}

We examine which perceptual information the small policy model requires and how ID-only adaptation affects the segmentation used to construct its observations. Additional evaluations compare visual interfaces and assess a second policy backbone.

\paragraph{Robot/gripper information}
We evaluate whether including the robot/gripper alongside the target object in the visual interface improves action learning. Table~\ref{tab:robot_mask_ablation} compares FMP policies trained with target-only observations and observations containing both the target and robot/gripper.

With target-only observations, success is 7.3\% on RoboMimic Lift and 5.3\% on RLBench CloseGrill. Including the robot/gripper increases success to 98.7\% and 74.7\%, respectively. The robot mask preserves visual information about the robot's position relative to the target, which may help the policy determine the required action. These results support retaining both entities when constructing observations for the small policy model in these tasks.

\begin{table}[t]
\centering
\small
\setlength{\tabcolsep}{6pt}
\renewcommand{\arraystretch}{1.15}
\caption{Robot/gripper mask ablation.
FMP success rates (\%), averaged over three seeds with
50 rollouts per seed.}
\label{tab:robot_mask_ablation}
\begin{tabular}{l|cc}
\hline
Task & Target-only & Target+Robot \\
\hline
RoboMimic Lift & 7.3 & 98.7 \\
RLBench CloseGrill & 5.3 & 74.7 \\
\hline
\end{tabular}
\vspace{-0.6em}
\end{table}

\paragraph{Segmentation model adaptation} We evaluate the effect of ID-only LoRA adaptation on the segmentation masks supplied to the visual interface. On ToiletSeatDown, we compare pretrained and adapted SAM3 under the ID setting and two held-out tabletop appearances. Table~\ref{tab:sam3_adaptation} presents an OOD segmentation example alongside mask IoU measurements for the robot/gripper and target object.

\begin{table}[t]
  \centering
  \caption{SAM3 before and after ID-only LoRA adaptation.
  Left: an OOD segmentation example.
  Right: mask IoU (\%) under ID and held-out appearances.}
  \label{tab:sam3_adaptation}

  \begin{minipage}[c]{0.26\linewidth}
    \centering
    \footnotesize

    Pretrained\par
    \vspace{2pt}
    \includegraphics[width=0.85\linewidth]
      {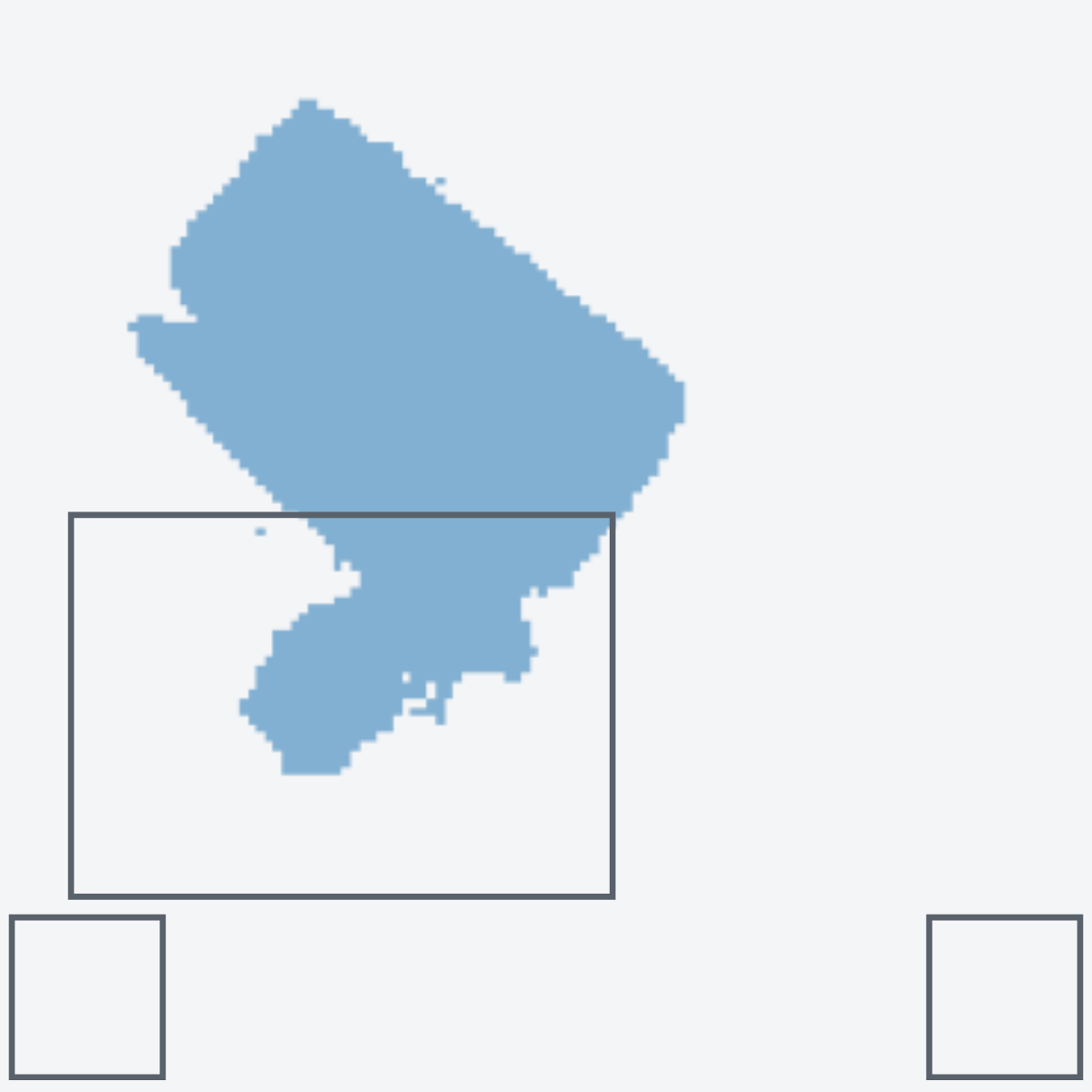}\par

    \vspace{4pt}

    LoRA-adapted\par
    \vspace{2pt}
    \includegraphics[width=0.85\linewidth]
      {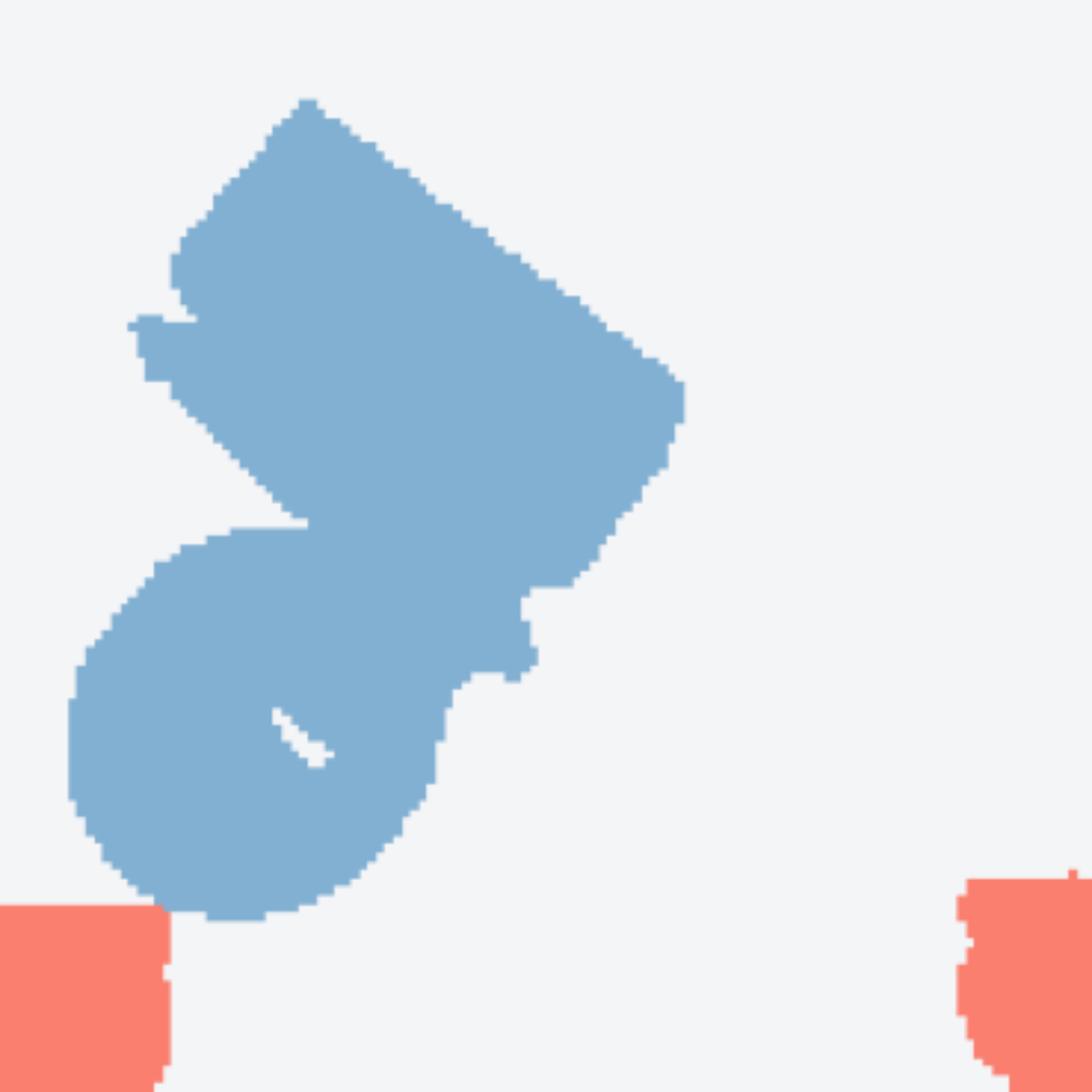}\par
  \end{minipage}%
  \hfill
  \begin{minipage}[c]{0.70\linewidth}
    \centering
    \footnotesize
    \setlength{\tabcolsep}{2pt}
    \renewcommand{\arraystretch}{1.30}

    \begin{tabular*}{\linewidth}
      {@{\extracolsep{\fill}}lcc@{}}
      \toprule
      Setting & Pretrained & LoRA-adapted \\
      \midrule

      \multicolumn{3}{c}{Robot/gripper IoU (\%)} \\
      \addlinespace[2pt]
      ID   & 0.0 & 99.3 \\
      OOD1 & 0.0 & 98.9 \\
      OOD2 & 0.0 & 99.1 \\

      \midrule
      \multicolumn{3}{c}{Object IoU (\%)} \\
      \addlinespace[2pt]
      ID   & 23.9 & 99.5 \\
      OOD1 & 32.9 & 99.2 \\
      OOD2 & 14.0 & 99.2 \\
      \bottomrule
    \end{tabular*}
  \end{minipage}

  \par\vspace{4pt}

  {\footnotesize
    \textcolor{samtarget}{\rule{0.75em}{0.75em}}
    \hspace{3pt}Target
    \qquad
    \textcolor{samrobot}{\rule{0.75em}{0.75em}}
    \hspace{3pt}Robot/gripper
    \par\vspace{2pt}
    Boxes highlight the incomplete target mask
    and missing gripper.
    \par
  }
\end{table}






Pretrained SAM3 obtains 0.0\% robot/gripper IoU and
14.0--32.9\% object IoU across the evaluated conditions.
LoRA adaptation increases these values to 98.9--99.3\%
and 99.2--99.5\%, respectively.
The qualitative example shows the recovery of the gripper
and more complete target-object masks after adaptation.

These results show that ID-only adaptation improves segmentation quality under both ID and held-out OOD appearances.
This supports preparing the perception component before
policy learning so that it can provide task-relevant masks
while remaining fixed during subsequent training and deployment.

\paragraph{Perception--Policy Interface}
We examine the visual interface between foundation-model perception and small policy models. We compare L0 with an ARRO-inspired masked-RGB baseline~\cite{zhu2025arro} on RoboMimic Lift and RLBench CloseMicrowave under object-color shifts. Within each task, both methods use the same SAM3 masks and FMP training and evaluation protocols. The masked-RGB baseline retains the original robot and target RGB values on the same constant background. Across the three object-color OOD conditions on Lift, mean success is 75.5\% for masked RGB and 89.5\% for L0. On CloseMicrowave, L0 achieves 85.3\% success, compared with 12.7\% for masked RGB. Across both tasks, these results support the value of fixed role-color rendering beyond foreground filtering in the perception--policy interface.

We further compare L1 with an alternative depth encoding.
Inspired by S$^{2}$-Diffusion~\cite{s2diffusion}, we implement
an FMP-based baseline, termed S2-style, that concatenates
one binary-mask channel and one normalized-depth channel.
A $1 \times 1$ convolutional adapter maps these channels
to the three-channel input expected by the unchanged
ResNet encoder.

\begin{table}[t]
\centering
\small
\setlength{\tabcolsep}{5pt}
\caption{Observation encoding comparison with FMP.
Success rates (\%) for S2-style encoding and L1.}
\label{tab:s2style_compare}
\begin{tabular}{llcccc}
\toprule
Task & Method & ID & OOD1 & OOD2 & OOD3 \\
\midrule
\multirow{2}{*}{CloseGrill}
& S2-style & 48.7 & 45.3 & 40.0 & 42.7 \\
& L1       & 82.7 & 81.3 & 82.0 & 82.0 \\
\midrule
\multirow{2}{*}{CloseMicrowave}
& S2-style & 64.0 & 60.7 & 58.7 & 61.3 \\
& L1       & 96.7 & 92.0 & 91.3 & 89.3 \\
\bottomrule
\end{tabular}
\vspace{2pt}
\end{table}

Table~\ref{tab:s2style_compare} shows higher ID and OOD
success with L1 on both tasks.
These results favor the proposed repainting interface
over the tested alternative encoding for connecting
foundation-model perception with action learning.

\paragraph{Evaluation with SmolVLA}
\label{sec:smolvla}
We additionally evaluate SmolVLA~\cite{shukor2025smolvla},
a compact pretrained vision--language--action model with
approximately 450 million parameters, on RLBench.
This evaluation examines whether the observation interface
also supports a compact VLA backbone under appearance shifts.
For each observation variant, we fine-tune and evaluate
SmolVLA using the corresponding inputs while retaining
its architecture.
Table~\ref{tab:rlbench_smolvla_obs} reports success under
the training tabletop appearance and two held-out colors.

\begin{table}[t]
\centering
 \tiny
\setlength{\tabcolsep}{4.0pt}
\renewcommand{\arraystretch}{1.15}
\caption{SmolVLA success rates (\%) under tabletop
color shifts on RLBench.}
\label{tab:rlbench_smolvla_obs}
\resizebox{\linewidth}{!}{
\begin{tabular}{l|ccc|ccc}
\hline
& \multicolumn{3}{c|}{CloseGrill}
& \multicolumn{3}{c}{CloseMicrowave} \\
\cline{2-7}
Obs. & ID & OOD1 & OOD2 & ID & OOD1 & OOD2 \\
\hline
RGB & 76.0 & 46.0 & 42.0 & 64.0 & 4.0 & 8.0 \\
L0  & 70.0 & 64.0 & 66.0 & 70.0 & 64.0 & 60.0 \\
L1  & 72.0 & 66.0 & 66.0 & 84.0 & 78.0 & 76.0 \\
\hline
\end{tabular}
}
\vspace{-0.6em}
\end{table}

The most pronounced gains occur on CloseMicrowave, where OOD success increases from 4.0--8.0\% with RGB to 76.0--78.0\% with L1. On CloseGrill, both observation variants improve OOD success but yield lower ID success than RGB, and adding depth provides a smaller benefit.

These results show that task-relevant perception from foundation models can also support a compact pretrained VLA, while the value of additional geometric information remains task-dependent. Together with the FMP evaluations, they provide evidence for foundation--policy coordination across the two tested backbones.


\section{Conclusion}
In this work, we presented a framework that coordinates pretrained vision foundation models with small policy models for robust visuomotor learning. Specifically, we designed an explicit perception--policy interface that constructs task-relevant observations through semantic repainting, with an alternative representation incorporating normalized target-region depth. Perception adaptation and policy learning are performed separately, enabling small policy models to learn actions from these observations without architectural changes. Experiments with FMP in simulation and on a real robot, together with SmolVLA evaluations on RLBench and component ablations, demonstrate the effectiveness of this coordination in improving robustness under appearance shifts. Future work will explore task-adaptive coordination to better align foundation-model perception with the action-learning requirements of small policy models.


\bibliographystyle{IEEEtran}
\bibliography{literature}

\end{document}